\documentclass[letterpaper, 10 pt, conference]{ieeeconf}  

\IEEEoverridecommandlockouts                              

\usepackage{graphicx} 

\usepackage{tabularx} 
\usepackage{booktabs}
\usepackage{hyperref}
\usepackage{multicol}
\usepackage[caption=false, font=normalsize, labelfont=sf, textfont=sf]{subfig}

\usepackage{algorithm,algorithmic}
\usepackage{amsmath}
\usepackage{tikz}
\usepackage{tikz-3dplot}
\usepackage{arydshln}
\usepackage{float}
\usepackage{amsfonts}
\usepackage{graphicx,  xcolor, psfrag, soul}
\usetikzlibrary{positioning, calc, patterns, angles, quotes}
\usetikzlibrary{quotes,angles}
\usepackage{subfig}
\usepackage{siunitx}
\usepackage[space]{cite}
\usepackage[caption=false]{subfig}

\newcommand{\fref}[1]{Figure~\ref{#1}}

\newcommand{\sref}[1]{Section~\ref{#1}}

\usepackage{xcolor}

\title{\LARGE \bf The WayHome: Long-term Motion Prediction on Dynamically Scaled Grids}
\author{Kay Scheerer$^{1}$, Thomas Michalke$^{2}$, J/"urgen Mathes$^{2}$
\thanks{$^{1}${\tt\small ScheererKayAr@gmail.com}}
\thanks{$^{2}$Robert Bosch GmbH,
Robert-Bosch-Platz 1, 
70839 Gerlingen-Schillerh\"ohe, Germany
        {\tt\small  Thomas.Michalke@de.bosch.com}}%
}
\begin{document}
\maketitle
\thispagestyle{empty}
\pagestyle{empty}
%
\begin{abstract}
One of the key challenges for autonomous vehicles is the ability to accurately predict the motion of other objects in the surrounding environment, such as pedestrians or other vehicles.
In this contribution, a novel motion forecasting approach for autonomous vehicles is developed, inspired by the work of Gilles et al.\cite{gilles2021HOME}. %
We predict multiple heatmaps with a neural-network-based model for every traffic participant in the vicinity of the autonomous vehicle; %
with one heatmap per timestep. The heatmaps are used as input to a novel sampling algorithm that extracts coordinates corresponding to the most likely future positions. %
We experiment with different encoders and decoders, as well as a comparison of two loss functions. %
Additionally, a new grid-scaling technique is introduced, showing further improved performance. %
Overall, our approach improves state-of-the-art miss rate performance for the function-relevant prediction interval of 3 seconds while being competitive in longer prediction intervals (up to eight seconds). %
The evaluation is done on the public \href{https://waymo.com/open/challenges/2022/motion-prediction/}{2022 Waymo motion challenge}. %

%
\end{abstract}
%

\section{Introduction}  
The term \textit{autonomous vehicle} refers to the ability of a vehicle to navigate and move without direct human intervention. This technology has the potential to revolutionize transportation by improving safety, reducing congestion, and increasing efficiency. However, there are still significant technical, legal, and ethical challenges that must be overcome to fully realize the potential of autonomous driving. The challenges include developing algorithms for perception, decision-making, and control, designing sensor systems and hardware architectures, evaluating performance and safety, or studying the social and economic impact of autonomous driving.
This work focuses on the task of motion prediction of other traffic participants and predicts up to eight seconds into the future based on an AI model. %
The motion prediction task must take several factors into account, such as traffic patterns, and road infrastructure, and should be accurate for a time horizon of several seconds. The prediction must be made in real time because the autonomous system receives new inputs from the environment and must quickly adapt to a changing environment. %
Hence, the autonomous system also needs to handle a large volume of data, should adapt to changing conditions, and has to be robust to noise and uncertainty in the data\cite{Wilson2021Argoverse2}. %
%
%
\begin{figure}[!t]
    \centering
\captionsetup[subfigure]{justification=centering}
        \subfloat[Input data\label{fig:topview}]{%
             \fbox{\includegraphics[width=0.145\textwidth]{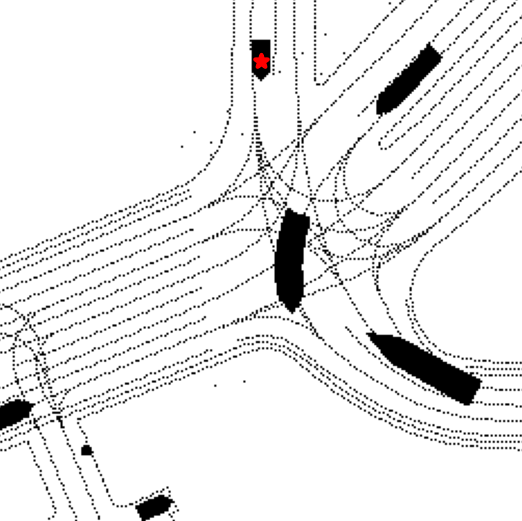}}}%
       \subfloat[Heatmap at $t=8$s\label{fig:heatmap8s}]{%
            \fbox{\includegraphics[width=0.145\textwidth]{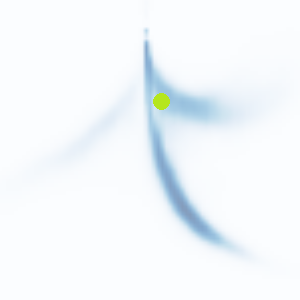}}}\hspace{0.1em}%
       \subfloat[Sampled coordinates\label{fig:samplingRunningExample}]{%
             \fbox{\includegraphics[width=0.145\textwidth]{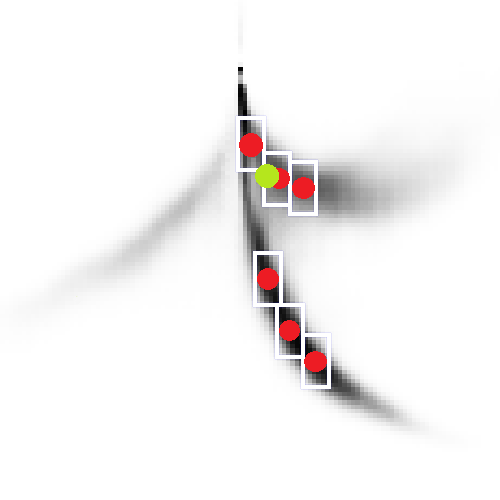}}}\hspace{0.1em}%
   \caption[Examples for ground-truth, input, and output]{Input data, heatmap outputs, and heatmap with sampled points. The target agent to predict is marked with a red dot in the input data. The ground-truth position at $t$ is marked in green.}
   \label{fig:runningExample}
\end{figure}
Our approach reuses some design decisions of Gilles et al.\cite{gilles2021HOME}: %
The space is discretized with a grid and is used to predict positions, where the value of a grid cell corresponds to the probability of the target agent to predict being in that grid cell. %
Multiple grids are predicted, each corresponding to a specific timestep $t$. %
The grid is denoted as a heatmap by Gilles et al.\cite{gilles2021HOME}. %
The grids are then used as the input for a sampling algorithm, that determines the most likely positions, restricting the output to a set of $n$ sampled positions, similar to public competitions in the motion prediction field\cite{waymoDataset, Ming2020Argoverse, Wilson2021Argoverse2}. %
The heatmaps are created with a neural network model inspired by the work of Gilles et al.\cite{gilles2021HOME}. %
Two inputs are given to the model, a discretized top-view image of the scene and sequences of features from the agents in the scene.
The abstract top view input is shown in \fref{fig:topview}, an example for a heatmap at $t_{8s}$ is shown in \fref{fig:heatmap8s} and six sampled points are visualized in red with the ground truth position in green in Figure~\autoref{fig:samplingRunningExample}. Samples are required in order to compare the prediction performance against other state-of-the-art approaches. %

The remaining paper is structured as follows:
\sref{chap:related_work} discusses related contributions and derives the still unresolved research questions. \sref{chap:method} gives a detailed description of our system-driven approach for motion prediction with heatmaps. %
In \sref{chap:Experiments}, we apply the approach to the public Waymo motion dataset and evaluate the performance against the public leaderboard.
%

%
\begin{figure*}[!ht]
    \centering
    \includegraphics[width=1\textwidth]{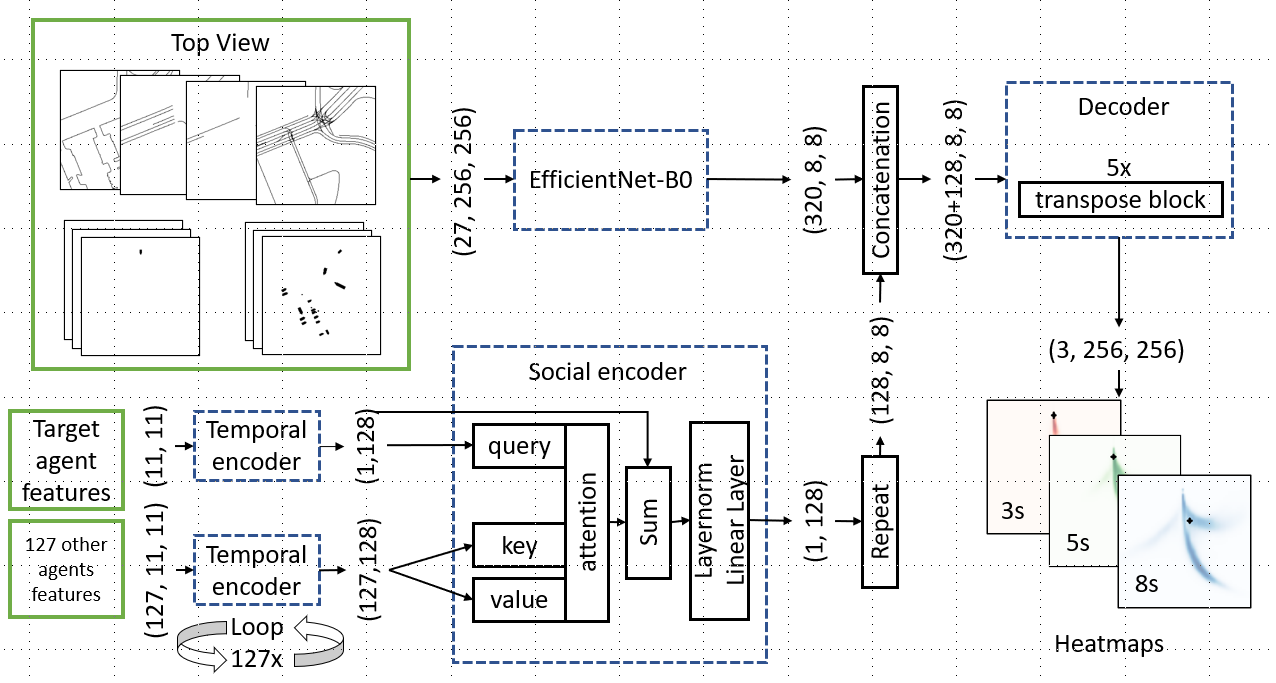}
    \caption[Architecture]{Architecture used in our approach. Multiple top-view encoders and heatmap decoders have been considered in our work. The input is shown in green boxes, the blue boxes are the different modules, and the arrows show the data flow.}
    \label{fig:WayHomeArchitecture}
\end{figure*}
\section{Related Work}     \label{chap:related_work}
\begin{table*}[bt!]
\centering
\begin{tabularx}{1.0\textwidth}{l c c l l l l l}   
      \toprule[1.25pt]
    {Name}& {Ref.} & {Year}  & {Agent repr.} & {Road repr.} & {Encoders} & {Output repr.}\\   
      \midrule[1.25pt]
    {MTRA} &\cite{MTRA2022Shi} %
    &2022  & polyline & polyline & Transformer & GMM\\
    {Wayformer}&\cite{Nayakanti2022wayformer}%
    &2022 & polyline & polyline &  Transformer & GMM\\
    {golfer}&\cite{tang2022golfer}%
    &2022 & polyline & polyline &  Transformer-like & GMM\\

    {StopNet}&\cite{kim2022stopnet}%
    &2022 & grid & grid & Point-Pillars\cite{lang2019pointpillars} and ResNet\cite{Resnet} & Occupancy grid  and  GMM\\
    {Mahjourian}&\cite{flowfieldsMahjourian}%
    &2022& grid & grid & Point-Pillars\cite{lang2019pointpillars} and MLP  & Occupancy grid \\
    {HOPE}&\cite{hu2022hope} %
    &2022 & grid & grid & Convolutions and  Swin-Transformer & Occupancy grid\\
    {VectorFlow}&\cite{huang2022vectorflow} & 2022 & grid + polyline & grid + polyline &  VGG-16+VectorNet & Occupancy Grid %
    \\
    {HOME}&\cite{gilles2021HOME}%
    &2021 & grid + polyline & grid & Convolutions, GRU and Attention &  Heatmap and Trajectories\\
    {GOHOME}&\cite{gilles2022gohome}%
    &2022 & polyline + graph & polyline + graphs  & Convolutions  and GRU  &Heatmap and Trajectories\\
    {HBEns}&\cite{Qian2022hbens}%
    &2022 & polyline + grid & polyline + grid &  HOME\cite{gilles2021HOME} and Multipath++\cite{Varadarajan2022multipath}  &  Heatmaps and Trajectories\\
      \midrule[1.25pt]
      {\textbf{WayHome (ours)}}&%
    &2023 & polyline + grid & grid &  Convolutions, GRU and Attention &  Heatmaps and Trajectories\\
    \bottomrule[1.25pt]
\end{tabularx}
\caption[Related work]{\label{table:related_work} Design decisions for recent approaches to motion forecasting, sorted by the output representations.}
\end{table*}
In the following, the known literature in the area of motion prediction is summarized. %

%

\subsection{AI Architectures}
The recent approaches in motion prediction are categorized into the representation of agents and roads as input and output representation, as well as the encoders used. %
%
A common approach to represent road elements or agents is as a set of points, so-called polylines\cite{gilles2022gohome}. %
The Transformer architecture presented by Vaswani et al.\cite{vaswani2017attention} can be used for encoding polyline representation and is the current state-of-the-art in the Waymo motion challenge\cite{MTRA2022Shi, Nayakanti2022wayformer, tang2022golfer}. %
Chai et al.\cite{Chai2019Multipath} introduce the Gaussian mixture model (GMM) to the field of motion prediction and show that the GMM output is advantageous for motion prediction. %
The GMM output has been adopted by various approaches\cite{Varadarajan2022multipath, Nayakanti2022wayformer, MTRA2022Shi, tang2022golfer,kim2022stopnet}. %
In the Waymo occupancy grid and flow field challenge, the future occupancy of agents in the scene needs to be predicted. %
To predict future occupancy one also has to predict future movement correctly, therefore, we also include approaches with occupancy grid output in the literature study. %
A common input representation of the occupancy grid challenge is a metric grid that also discretizes the space, %
like an abstract top-view image of the scene (also called bird's eye view)\cite{hu2022hope,flowfieldsMahjourian}. %
However, polylines can also be used\cite{huang2022vectorflow}. %
Notably, there exist combinations of coordinate and grid output. %
Kim et al.\cite{kim2022stopnet} predict an occupancy grid and a GMM in parallel with separate decoders. %
Gilles et al.\cite{gilles2021HOME, gilles2022gohome} predict a grid with probabilities for the future position at $t_{3s}$, the so-called heatmap. %
Furthermore, Gilles et al. generate trajectories by sampling likely non-redundant coordinates from the heatmap. The coordinates serve as input to another model together with the target agent's past feature states. %
The model outputs multiple trajectories, each with one of the sampled coordinates as the endpoint. %
In more detail, the sampled coordinates are non-redundant because the coordinates always have at least a distance $D$ between them\cite{gilles2021HOME}, in other words, the same coordinate is not predicted multiple times. %
We argue, that it is not necessary to generate trajectories, as the trajectories might be redundant for earlier timesteps again, and it is more beneficial to predict only the waypoints, incoherent over time, such that they are non-redundant for each timestep. %
%
%
Nayakanti et al.\cite{Nayakanti2022wayformer}, Shi et al.\cite{MTRA2022Shi}, Varadarajan et al.\cite{Varadarajan2022multipath} also follow a refinement strategy to pick non-redundant coordinates by first predicting a larger number of coordinates per timestep and then pick coordinates by non-maximum-suppression\cite{MTRA2022Shi} or by a k-means-like algorithm\cite{Varadarajan2022multipath, Nayakanti2022wayformer}. %
Determining the best distance that should be between two coordinates depends on the task at hand. %
One can base the distance on the threshold used in the miss rate metric to optimize the miss rate score, as done by Gilles et al.\cite{gilles2022gohome, gilles2021HOME}. %


\subsection{Metrics}\label{section:metrics}
Common metrics used to evaluate coordinate output are presented in the following. %
Commonly, multiple coordinates per timestep are predicted, and the closest prediction is used for the metric score.
Multiple competitions restrict the number of predictions, for example, to six in the Waymo motion competition\cite{waymoDataset} or both the Argoverse challenges\cite{Ming2020Argoverse, Wilson2021Argoverse2}.
\paragraph{Minimum displacement error}
The minimum displacement error is the distance of the closest predicted coordinate at timestep $t$ to the ground truth at $t$. The minimum average displacement error is the minimum of all averages over the timesteps, and the minimum final displacement error is the error for the last timestep\cite{waymoDataset}. %
A lower displacement error is better. %
\paragraph{Miss Rate}
\begin{table}[ht]
\centering
\begin{tabular}{|S|SS|} \hline
    {t} & {$\lambda^{lat}_t$} & {$\lambda^{lon}_t$}\\ \hline
    {3s}  & 1.0m & 2.0m \\
    {5s}  & 1.8m & 3.6m \\
    {8s}  & 3.0m & 6.0m \\\hline
\end{tabular}
\caption[Miss-rate thresholds]{\label{tab:waymoLambda} Values for $\lambda_t$, Waymo miss-rate metric}
\end{table}
The miss rate is the proportion of predicted agents for which none of the predicted coordinates at timestep $t$ is inside a threshold area of the ground truth at $t$. %
The threshold area for both the Argoverse competitions\cite{Ming2020Argoverse, Wilson2021Argoverse2} is a 2m circle around the ground truth position. %
The Waymo motion challenge uses a rectangular dynamic threshold $\lambda_t$ that is twice as long in the longitudinal direction than in the latitudinal direction. %
The target agent is rotated such that the longitudinal direction is always in direction of the heading at the future timestep $t$. %
The threshold $\lambda_t\in\mathbb{R}^2$ is shown in \autoref{tab:waymoLambda}, which is furthermore scaled by $\gamma(v)\in[0.5, 1]$, based on the velocity $v$ of the target agent at the current timestep $t_{0s}$\cite{waymoDataset} (see \autoref{eq:waymo_gamma}). %
A lower miss rate is better. %

\begin{equation}
\begin{aligned}\label{eq:waymo_gamma}
  \gamma(v)  &=\frac{f(v)}{2}+0.5 \\
    f(v) &= \max(\min(h(v),1),0) \\
    h(v) &= \frac{v - 1.4\frac{m}{s}}{11\frac{m}{s} - 1.4\frac{m}{s}}  
\end{aligned}
\end{equation}
\paragraph{Soft mean average precision}
The miss rate is used as a binary indication in the calculation of the Soft mean average precision (Soft mAP or SmAP). %
From the non-missed predictions, the one with the largest confidence is picked and defined as 0 or 1 based on a confidence score threshold $c$. %
The area under the precision-recall curve with various confidence score thresholds $c$ is the average precision\cite{waymoDataset}. %
Moreover, the future trajectories are partitioned into buckets, like trajectories going to the left, the right, or straight ahead. %
The average precision is calculated per bucket and averaged over all buckets is the mean average precision\cite{waymoDataset}.
A higher Soft mAP is better. %


\subsection{Grid Size}
\begin{figure}[htb]
    \centering
    \subfloat[2 pixels per meter\label{fig:resize2}]{%
        \fbox{\includegraphics[width=0.225\textwidth]{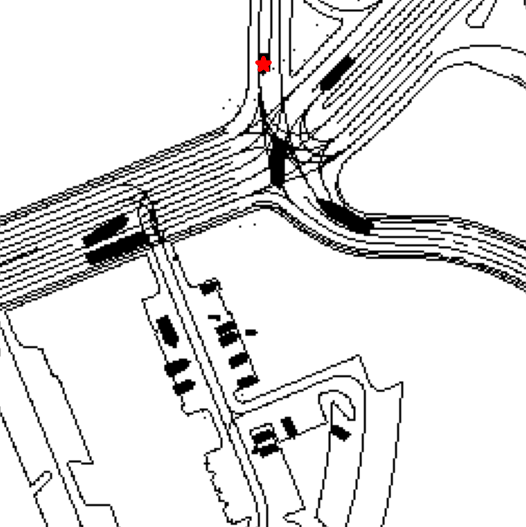}}}\hspace{0.1em}%
    \subfloat[4 pixels per meter\label{fig:resize4}]{%
        \fbox{\includegraphics[width=0.225\textwidth]{pictures/resize_4.png}}}\hspace{0.1em}%
    \caption[Grid scale difference]{Grid scales at resolution $256\times256$ }
    \label{fig:differentResizes}
\end{figure}
Top view grids are common choices to represent road structure or agents \cite{flowfieldsMahjourian, gilles2022gohome, gilles2021HOME, kim2022stopnet}. %
The grid size is determined by two hyperparameters: the resolution (or the number of pixels) and the pixels per meter\footnote{or the inverse of it, the $\frac{meter}{pixel}$}, that is, how much physical space each pixel corresponds to, for example, each pixel corresponds to $0.25m^2$ or $1km^2$. %
Two examples of a grid with the same resolution $256\times 256$ and a different physical resolution are shown in \autoref{fig:differentResizes}. %
The size of the grid depends on the task at hand, e.g., if the movement of ships is predicted a grid could be needed that covers the whole world. In our use case, a grid with an area of $256m^2$ is sufficient to predict vehicle motions for the next eight seconds. %
However, the grid size is often just stated\cite{Chai2019Multipath, kim2022stopnet, gilles2022gohome, gilles2021HOME}. %
There are some requirements for the grid size, including that it captures most of the scene and that is precise in its representation. %
When predicting future positions with a grid, it is important to have a grid that is large enough otherwise the agents might be outside of the grid bounds and cannot be predicted with the grid. %
For example, an agent starts somewhere in a grid that spans a region of $100m^2$ and drives $300m$ in a straight direction which is then outside of the grid boundaries. %
\autoref{Tab:gridsizes} shows different design choices for the grid size parameters of recent approaches from the literature. %
Different from state-of-the-art we proposed a dynamic grid scaling, adopt the AI architecture accordingly, and prove the benefits by evaluating the approach on the Waymo motion dataset.
\begin{table}[!ht]  
  \begin{center}
    \begin{tabular}{lllSc}
      \toprule[1.25pt]
      Method    & Year & Reference &$\frac{meter}{pixel}$ & resolution \\ \midrule
        HOME & 2021 & \cite{gilles2021HOME} & 0.5 & $288\times 288$ 
      \\
        GOHOME & 2022 & \cite{gilles2022gohome} & 0.5 & $384\times 384$ 
      \\
        StopNet  & 2022 & \cite{kim2022stopnet} & 0.4 & $400 \times 400$
      \\
      
        Mahjourian et al. & 2022 & \cite{flowfieldsMahjourian} & 0.2 & $400 \times 400$
        \\
        HOPE &2022&\cite{hu2022hope} & 0.16 & $768 \times 768$
      \\
      VectorFlow & 2022 &\cite{huang2022vectorflow} & 0.31 & $256\times 256$
      \\
      HBEns & 2022 & \cite{Qian2022hbens} & {not given} & $320\times 320$
      \\
      \midrule
      \textbf{Ours} \\
      \midrule
      \textbf{Velocity scaling}  & &   & \textbf{0.5 --- 1} & \textbf{256$\times$ 256} \\
      \textbf{Time scaling}  & &   & \textbf{0.3 --- 1} & \textbf{256$\times$ 256} \\
      \bottomrule[1.25pt]
    \end{tabular}
  \end{center}
  \caption{Selected state-of-the-art approaches and the (rounded) choices for $\frac{{meter}}{{pixel}}$ and resolution}
  \label{Tab:gridsizes}
\end{table}

\subsection{Loss Functions}
Both the focal loss and the cross-entropy loss can be used for probabilistic grid output, by calculating the loss per grid cell and averaging over all cells\cite{gilles2022gohome, gilles2021HOME, flowfieldsMahjourian}. %
The focal loss is used in the work of Gilles et al.\cite{gilles2022gohome, gilles2021HOME}; the cross entropy is used in multiple other works\cite{kim2022stopnet, flowfieldsMahjourian, huang2022vectorflow}. %
However, there is no comparison to other established loss functions, like the cross-entropy function. %

Our work contributes the following aspects:
\begin{itemize}
	\item novel dynamic grid scaling,
	\item a miss rate comparison of the focal loss and the cross-entropy,
	\item adapting the work of Gilles et al.\cite{gilles2021HOME} to the Waymo motion metrics
\end{itemize}

Major advantages are:
\begin{itemize}
        \item 1. dynamic physical input and output resolution based on the scenario,
        \item 2. optimized AI architecture design based on HOME\cite{gilles2021HOME}.
\end{itemize}

\section{Method}             \label{chap:method}
The following section describes the proposed prediction method in detail, especially elaborating on the network architecture to predict heatmaps, the sampling procedure to retrieve coordinates from the predicted heatmaps, and finally a novel grid scaling technique.
%
%
%

\subsection{Network Architecture}

Our architecture is inspired by the work of Gilles et al.\cite{gilles2021HOME} and is shown in \autoref{fig:WayHomeArchitecture}. %
The HOME architecture is an encoder-decoder structure, with separate encoders receiving different input representations. %
The inputs to the model are a top-view for the \textit{top-view encoder}, and agent features for two separate \textit{temporal encoders}. %
The output of the \textit{temporal encoders} is the input to a so-called \textit{social encoder}. %
The output of \textit{top-view encoder} and the \textit{social encoder} are concatenated and then decoded to the heatmaps. %
The task of the temporal encoder is to create a latent representation to be used in the social encoder, which encodes the social interaction between agents. %
The two input representations are a top view of the scene and a sequence of past agent states, further, split into the target agent and all other agents. %
Our architecture is built modular, such that the top-view encoder and the decoder can be replaced easily. %
Only the input and output representations are fixed. %
However, some parts of the architecture are replaced. %
Therefore, we will name our implementation \textit{WayHome}, to prevent confusion between the approach of Gilles et al.\cite{gilles2021HOME} and our approach. %
\paragraph{Agent Features + Encoder}
The agent features are split into two groups, the target agent to predict, and all other agents. %
Each group uses one encoder. %
There are a maximum of 128 agents in our dataset, therefore, the other agents are 127. %
The agent features for each timestep are: the x and y position, the valid flag, the velocity in the x and y direction, the velocity vector magnitude, the width and length, the yaw angle, the velocity vector yaw angle, and the agent type, encoded as a number, e.g. pedestrian=0 and vehicle=1. 
If there are fewer agents or the agents are invalid, the valid flag indicates so. %
The \textit{temporal encoder} in \fref{fig:WayHomeArchitecture} is a 1D convolution followed by batch norm and \textit{ReLU} activation function, increasing the feature shape from 11 to 64, followed by a Gated Recurrent Unit (GRU) that encodes the size 64 to size 128. %
The target agent is encoded with one \textit{temporal encoder}, and all other agents share the same \textit{temporal encoder}. %
We prune the time dimension from the \textit{temporal encoder} output and only use the last timestep, to enable using it as input to the social encoder. %
The social encoder is implemented as depicted in \fref{fig:WayHomeArchitecture}. %

\paragraph{Top-view + Encoder}
Our top-view input representation uses a separate image channel per road category, that is, lane centers, white lines, yellow lines, road edges, and crosswalks. %
The target agent is split from the other agents. %
Moreover, a separate top-view is used per timestep, that is, a total of eleven images for the target agent and another eleven images for all other agents for 11 timesteps for 1.1 seconds. %
Overall, our top-view image has $2\cdot 11+5$ channels. %
For the top-view encoder, we consider multiple choices that have shown great success in their respective research fields: the ResNet by He et al.\cite{He0216Resnet}, the EfficientNet by Tan and Le \cite{TanAndLe2019efficientnet}, and the Vision Transformer (ViT) by Dosovitskiy et al.\cite{Dosovitskiy2020ViT}. %
Zheng et al. used a ViT topped with a decoder to generate images for semantic segmentation, which inspired us to use the ViT to predict heatmaps. %
Moreover, we consider one of the decoders presented by Zheng et al.\cite{zheng2021setransformer}, the \textit{progressive UPsampling} decoder. In terms of configurations, we employ the ResNet-18, the EfficientNet-B0, and multiple ViT configurations because the default configuration presented by Dosovitskiy et al.\cite{Dosovitskiy2020ViT} with over 90 million parameters did not converge well in our training. We pick three configurations presented by Steiner et al.\cite{Dosovitskiy2020ViT}: the ViT with $\approx5$ million parameters with patch size 16, denoted ViT-5m/16 in the following, and the ViT with $\approx22$ million parameters once with patch size 16 and once with patch size 32, denoted ViT-22m/16 and ViT-22m/32 respectively. %
\paragraph{Decoders}
Two decoders are considered, the decoder by Gilles et al. which doubles the image resolution with every decoder block with transpose convolutions\cite{gilles2021HOME} and the \textit{progressive UPsampling} decoder by Zheng et al.from the field of semantic segmentation that doubles the image resolution in every layer with bi-linear interpolation\cite{zheng2021setransformer}. %
We denote the decoders after the module that doubles the image resolution in each layer, that is, \textit{transpose convolutions decoder} and \textit{interpolation decoder} respectively. %
After doubling the size, both decoders are followed by convolution, \textit{batch norm}, and \textit{ReLU}. %
The transpose convolution decoder halves the image channels in each step, while the interpolation decoder uses fixed 256 channels in each layer. %
Both decoders are topped with a convolution followed by a \textit{Sigmoid} activation, creating heatmaps with pixel values $\in [0,1]$. %

%

\subsection{Sampling Procedure}
\begin{figure}[htb]
    \captionsetup[subfigure]{justification=centering}
    \centering
    \subfloat[Sampled coordinates with approximated areas\label{fig:approxArea}]{%
        \fbox{\includegraphics[width=0.225\textwidth]{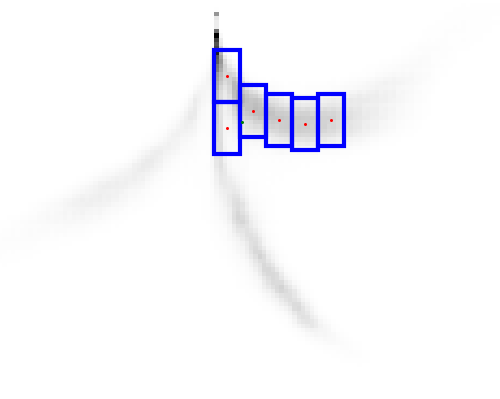}}}\hspace{0.1em}%
    \subfloat[Actual threshold areas of the sampled coordinates\label{fig:actualArea}]{%
        \fbox{\includegraphics[width=0.225\textwidth]{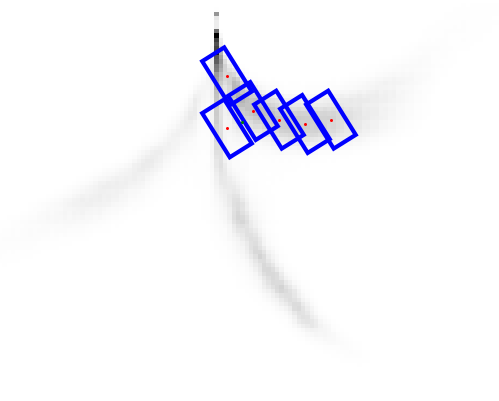}}}\hspace{0.1em}%
    \caption{Sampled coordinates in red and the space they cover as a blue rectangle around}
    \label{fig:sampledCoords}
\end{figure}
Our sampling algorithm is inspired by the sampling algorithm of Gilles et al.\cite{gilles2021HOME}. %
First, we use the threshold $D$ of Waymo's miss rate metric, that is, $D=\gamma(v)\cdot \lambda_t$ (see \sref{section:metrics}). %
$D$ corresponds to the threshold from the ground-truth position, in the latitudinal and the longitudinal direction. %
We turn around the idea and use a threshold around each predicted position, and if the ground-truth is inside a threshold of one of the predicted positions, the prediction is not missed. %
Therefore, we say that each predicted position covers an area around it, in which the ground-truth needs to be. %
The space should be covered as efficiently as possible, with only six predicted positions.\footnote{due to the restriction to six predictions for Waymo's miss rate} %
Overlapping areas are considered inefficient because the space is covered by multiple coordinates. %
Therefore, we chose the coordinates such that the covered space does not overlap. %
An example of sampled points from a heatmap is shown in \fref{fig:approxArea}. %
The minimum distance between two points, such that the areas do not overlap, is $2\cdot D$, and the threshold is doubled again, to correspond to the left and right, and up and down. %
Moreover, the threshold is multiplied with the $\frac{pixel}{meter}$, the corresponding threshold size in image space, which is then rounded down. %
The final size of the area covered is calculated with \autoref{eq:kernel_size}. %
\begin{equation}
size_{kernel} = \lfloor 4\cdot\gamma(v) \cdot \lambda_t \cdot x\frac{pixel}{meter} \rfloor
\label{eq:kernel_size}
\end{equation}
Following, a convolution with stride 1 and with kernel size equal to $size_{kernel}$ is used on the heatmap, summing up values in an area around each pixel. %
The coordinates are then sampled from the convolved heatmap, in a greedy fashion. %
The procedure is shown in Algorithm \ref{algo:sampling}. %
 \begin{algorithm}
 \caption{Sampling algorithm}
 \begin{algorithmic}[1]
 \label{algo:sampling}
 \renewcommand{\algorithmicrequire}{\textbf{Input:  }}
 \renewcommand{\algorithmicensure}{\textbf{Output:  }}
 \REQUIRE convolvedHeatmap, numSamples, threshold
 \ENSURE  coordinates, confidences
  \STATE coordinates, confidences = list()
  \FOR {$i = 0$ to $numSamples$}
    \STATE confidence = max(convolvedHeatmap)
    \STATE confidences.append(confidence)
    \STATE pixelIndex = argmax(convolvedHeatmap)
    \STATE globalCoord = calcGlobalCoord(pixelIndex)
    \STATE coordinates.append(globalCoord)
    \STATE indices = pixelsInTheArea(pixelIndex, threshold)
    \STATE convolvedHeatmap[indices] = $-\infty$
  \ENDFOR
 \RETURN coordinates, confidences
 \end{algorithmic} 
 \end{algorithm}
Basically, the maximum pixel is chosen, the value is chosen as the confidence for the sampled coordinate and the position is calculated into global coordinates. %
Moreover, all pixels in the vicinity of the chosen pixel are set to $-\infty$, corresponding to the area that this sampled coordinate covers, such that the pixels in the area cannot be sampled again. %
However, the $size_{kernel}$ is only an approximation to the actual miss rate threshold for two reasons, first, we discretize the threshold, and second, the actual threshold is calculated in an agent-centric coordinate system at $t$, where everything is rotated by the yaw angle of the target agent at $t$, which is unknown. %
Hence, the actual threshold is rotated, and therefore, the actual threshold areas can  overlap slightly. %
The approximated areas (blue rectangles) of the sampled coordinates (red dots) are shown in \fref{fig:approxArea}, and the actual areas are shown in \fref{fig:actualArea}. %
The space is not sampled as efficiently for left and right turns. %
In conclusion, the sampling procedure is adapted to the metrics to sample the best performing positions in terms of Waymo's miss rate and the predicted heatmap.


\subsection{Grid Scaling}
Top-view grids are both input and output of our approach. %
The area that a top-view covers affects performance because if the area is too small then the predictions might be outside of the grid and will be counted as a miss. %
Moreover, if the span area is large, each grid cell corresponds to more space which is imprecise. For example, with a resolution of $256\times 256$ and each pixel corresponds to a square meter the grid spans $256m^2$ but predictions can only be made with a precision of one square meter. %
By increasing the grid resolution the area of each pixel can be reduced, yet larger resolutions require more computation. %
Therefore, the region span by the grid is scaled such that it spans as much space as possible and the ground-truth positions are inside the grid area. %
Moreover, we keep the resolution fixed at $256\times 256$ for both input and output grids. %
Two dynamic scaling techniques are experimented with: one based on the timestep to predict, the other based on the velocity of the target agent at $t=0s$. %

\paragraph{Time-based scaling}
To find a good scaling value per timestep, three different scaling values $x\frac{pixel}{meter}, x\in\{1,2,3\}$ are used. %
For each $x$ a grid is created and we measure the proportion of ground-truths inside the grid boundaries. %
If the proportion falls below 99.9\%, the grid is considered too small. %
The values $x$ for which the grid is large enough are: $t_{3s}\rightarrow 3\frac{pixel}{meter}$, $t_{5s}\rightarrow 2\frac{pixel}{meter}$, and $t_{8s}\rightarrow 1\frac{pixel}{meter}$. %
For this scaling, only the three timesteps are defined. %
However, the idea can also be extended to other timesteps as well. %

\paragraph{Velocity-based scaling}
For the velocity-based scaling, we choose $x\frac{pixel}{meter}$ as $\frac{1}{\gamma(v)}\in[1,2]$, where $\gamma(v)$ is the velocity-based scaling from Waymo's miss rate metric. %
This scaling factor has the advantage, that the scaling of the kernel size in our sampling algorithm and the scaling of the threshold area are one. 

\section{Experiments}      \label{chap:Experiments}
\begin{table*}[!htp]
  \begin{center}
    \begin{tabularx}{1.0\textwidth}{ll|cccc|cccc|c}
      \toprule[1.25pt]
      {Encoder} &{Decoder}& {$\downarrow$ MR 3s} & { MR 5s} & { MR 8s} & { MR Avg} &$\uparrow$ SmAP 3s & SmAP 5s & SmAP 8s & SmAP Avg &\#Params \\   \hline
    &{\textit{4 decoder layers}}&&&&&&&&\\
      \hline
    {WH encoder}&{Transpose conv.}  & 9.34\%  & 12.41\% & 16.41\%  & 12.72\%&48.20\%& 39.54\%& 29.89\%& 39.21\% &4.3M \\ 
    {WH encoder}&{Interpolation}  & 9.71\%  & 12.34\% & 16.54\%  & 12.86\%&47.33\% & 40.74\% & 30.81\% & 39.63\% &5.1M \\ 
      \hline
    {ViT-5m/16}&{Transpose conv.}  & 9.13\%  &\textbf{ 11.70}\% &\textbf{15.29}\%  & \textbf{12.04}\%& 49.96\% & 42.80\% &\textbf{33.28\%}&42.01\% &8.2M \\
    {ViT-5m/16}&{ Interpolation}  & 9.20\%  & 12.02\% & 15.43\%  & 12.37\% & 49.02\% & 41.11\% & 33.04\%& 41.77\%& 9.5M\\
      \hline
    {ViT-22m/16}&{Transpose conv.}  & 9.20\%  & 11.81\% & 15.96\%  & 12.32\% & 49.98\%& 42.11\% & 32.12\%& 41.40\% & 25.8M\\ 
    {ViT-22m/16}&{Interpolation}  & 9.28\%  & 12.37\% & 16.18\%  & 12.61\% & 49.08\%& 41.18\%& 32.54\%& 40.93\%&27.2M\\ 
      \hline
    & {\textit{5 decoder layers}}&&&&&&&&\\
      \hline
    {EfficientNet}&{Transpose conv.}  &\textbf{8.69}\%   & 11.72\% & 15.76\%   & 12.06\% &\textbf{50.70\%}& \textbf{43.12\%}& 32.49\%& \textbf{42.10\% }&6.5M\\
    {EfficientNet }&{Interpolation}  & 9.94\%   & 12.52\% & 16.27\%   & 12.91\% & 47.25\% & 40.17\% & 30.93\% & 39.45\% & 7.3M  \\
      \hline
    {ResNet}&{Transpose conv.}  & 10.31\%  & 13.41\% & 17.52\%  & 13.75\% &46.30\%& 39.48\%& 29.93\%& 38.57\%& 14.0M\\ 
    {ResNet}&{Interpolation}  & 10.58\%  & 14.05\% & 18.48\%  & 14.37\% & 44.97\% & 38.47\% & 29.59\% & 37.68\%& 14.8M \\ 
      \hline
    {ViT-22m/32}&{Transpose conv.}  & 9.95\%  & 13.27\% & 17.19\%  & 13.47\% &46.58\%& 39.48\%& 29.93\%& 38.57\% &34.5M\\
    {ViT-22m/32}&{Interpolation}  & 10.33\%  & 13.90\% & 18.03\%  & 14.09\% &45.53\%& 38.85\% & 30.03\% & 38.14\%&35.2M\\
      \bottomrule[1.25pt]
    \end{tabularx}
  \end{center}
  \caption{Comparison of different top view encoders and decoders, in terms of miss rate (MR) and in terms of soft mean average precision (SmAP) 
  }
  \label{Tab:designChoiceComparison}
\end{table*}

In the following section, three experiments are conducted: first a comparison of the cross-entropy loss and the focal loss that has been used by Gilles et al.\cite{gilles2021HOME}. The comparison of loss functions is combined with the comparison of different grid sizes, to evaluate whether there are consecutive findings across different grid sizes. Second, architectural design choices are researched by exploring multiple choices for top-view encoders and decoders. %
Last, a comparison of our approach to state-of-the-art models is presented. %
%




\subsection{Grid Scaling}
\begin{figure}[!ht]
\centering
    \includegraphics[width=0.35\textwidth]{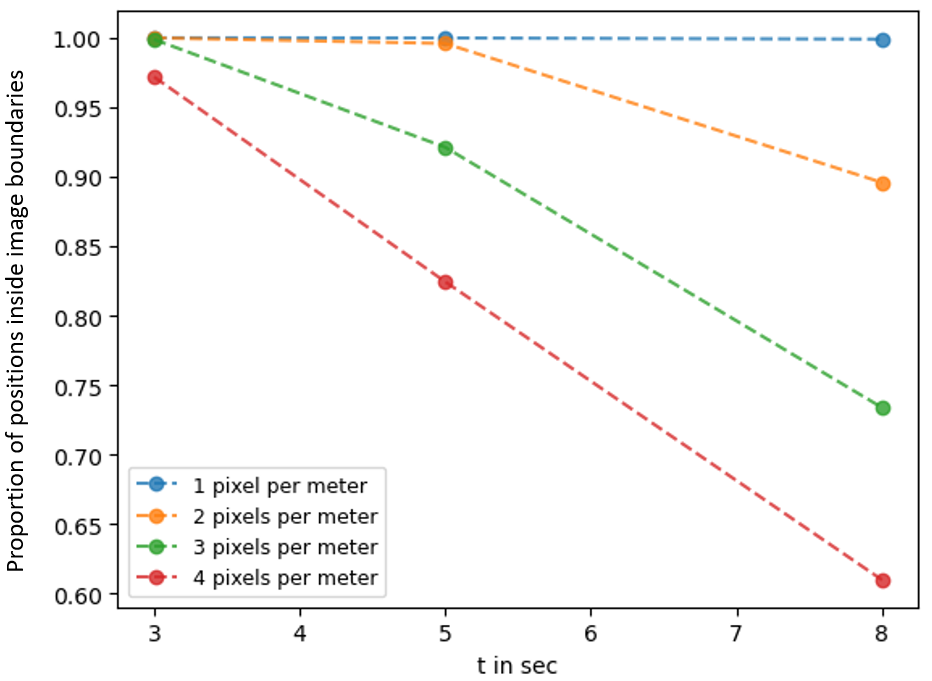}
    \caption{Proportion of ground truth positions inside the image boundaries $\frac{pixel}{meter}$  }
    \label{fig:sum_gt}
\end{figure}
The \autoref{fig:sum_gt} shows the proportion of ground truth positions inside of grid boundaries for different grid sizes. %
With a fixed grid size of resolution $256\times 256$, only $1\frac{pixel}{meter}$ is sufficient to capture all positions, and all other values have significant drops. %
The proportion of non-captured position is also the minimum value for the miss rate in our case because the position cannot be predicted if it is outside the grid boundaries. %
\begin{figure}[!ht]
\centering
    \includegraphics[width=0.35\textwidth]{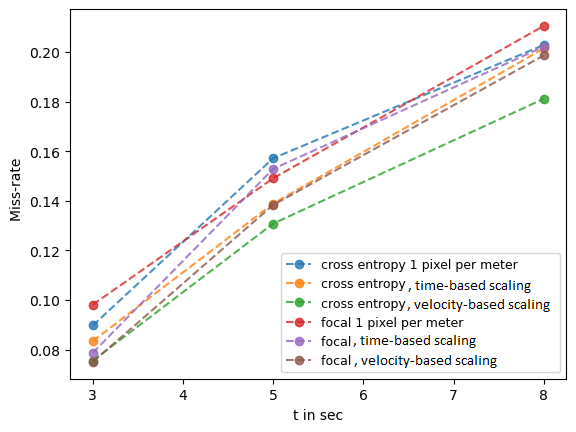}
    \caption{miss rate comparison for the dynamic scaling approaches versus a static grid size with 1 pixel per meter}
    \label{fig:mr_dynamicscale}
\end{figure}
In \autoref{fig:mr_dynamicscale} the miss rate scores for the time-based and velocity-based dynamic scaling approaches are compared using one static size of $1\frac{pixel}{meter}$. %
Moreover, the focal loss and cross-entropy loss are compared against each other, for a total of six models trained, each for exactly seven epochs on all \textit{tracks to predict} Waymo motion training dataset\footnote{approximately 1.6 million samples per epoch}. %
The miss rate scores are calculated on 20000 samples of the validation dataset for the timesteps $t_x, x\in\{3s,5s,8s\}$. %
The best-performing approach is the velocity-based scaling, for both loss functions on $t_x$. %
The cross-entropy loss is performing slightly better with velocity-based scaling than the focal loss. %
Compared to the static grid size, the dynamic grid sizes almost always perform better, except for the focal loss at $t_{5s}$. %

\subsection{Architectural Design Choices}
%
%
%
%
In \autoref{Tab:designChoiceComparison} the Miss Rate (MR) and Soft mean Average Precision (SmAP) scores are given for twelve model configurations, %
together with the number of parameters in millions.
The output size of some configurations is half the size, therefore, an additional decoder layer is used for those to create the same output image resolution. %
Higher scores are better for SmAP and lower scores are better for MR scores. %
The two best configurations are the ViT-5m/16 and the EfficientNet, both with the transpose convolution decoder; the configuration with the ViT-5m/16 reaches the best performance for the later timestep at $t_{8s}$ for both SmAP and MR, the configuration with the EfficientNet performs better for $t_{3s}$, on both MR and SmAP. %
For $t_{5s}$ the configuration with the EfficientNet is best in terms of SmAP and the ViT-5m/16 configurations is best in terms of MR. %
The performance difference between the two decoders is not large. %
The \textit{interpolation decoder} has worse performance in terms of MR, except the configuration with the \textit{WayHome} top view encoder at $t_{5s}$. %
In terms of average SmAP, the difference is in favor of the \textit{transpose convolution decoder}, except for the \textit{WayHome} and the ViT-22m/32 top view encoders. %
The largest difference in terms of average SmAP can be found together with the EfficientNet top view encoder with 2.65\%. %
\begin{table*}[!ht]
\centering
\begin{tabularx}{1.0\textwidth}{llSSSS|SSSS}  
      \toprule[1.25pt]
    {Name} & {Year} & {$\downarrow$ MR 3s} & {MR 5s} & {MR 8s}& {MR Avg} &{$\uparrow$ SmAP 3s} & {SmAP 5s} & {SmAP 8s} & {SmAP Avg}  \\   \hline
    {MTRA \cite{MTRA2022Shi}}  &2022     &10.07\%          & \textbf{11.46}\%   & \textbf{13.28}\%     & \textbf{11.60}\%%
    & \textbf{51.80}\% & \textbf{45.92}\% & \textbf{40.12}\% & \textbf{45.94}\%
    \\
    {\textbf{WayHome (Ours)}}    &2022     &\textbf{9.00}\%  &12.01\%              &15.58\%               & 12.20\% %
    & 49.57\%   & 42.58\% & 33.15\% & 41.77\%
    \\
    {Wayformer (multi-axis)\cite{Nayakanti2022wayformer}}& 2022 &9.46\%   &11.54\%              &15.86\%               & 12.28\%%
    & 50.76\%  & 44.07\% & 35.22\%  & 43.35\% 
    \\
    {Wayformer (factorized)\cite{Nayakanti2022wayformer}}&2022 &9.50\%   &11.58\%              &15.79\%               & 12.29\%%
    & 50.63\%  & 43.07\% & 34.10\%  & 42.60\%
    \\
    {MTR \cite{MTRA2022Shi}}   &2022      &11.77\%          &13.17\%             &15.59\%               & 13.51\%%
    & 48.39\%  & 41.98\% & 36.12\%  & 42.16\%
    \\
    {golfer\cite{tang2022golfer}}&2022    &9.62\%           &13.11\%              &17.88\%               & 13.54\% %
    & 50.34\%  & 43.15\% & 34.26\%  & 42.59\% 
    \\
    {HBEns\cite{Qian2022hbens}} &2022    &12.65\%           &14.86\%              &20.25\%               & 15.92\%%
     & 45.31\%  & 38.35\% & 30.25\%  & 37.97\%
    \\
    
      \bottomrule[1.25pt]
\end{tabularx}
\caption[State-of-the-Art comparison, MR]{\label{table:missrates_test}Miss rate (MR) scores and soft mean average precision (SmAP) from the 2022 \textit{Waymo motion} leaderboard as of 20.12.2022 
Sorted by average MR. 
} 
\end{table*}
\subsection{Sampling performance}
\begin{figure}[!ht]
\centering
    \includegraphics[width=0.35\textwidth]{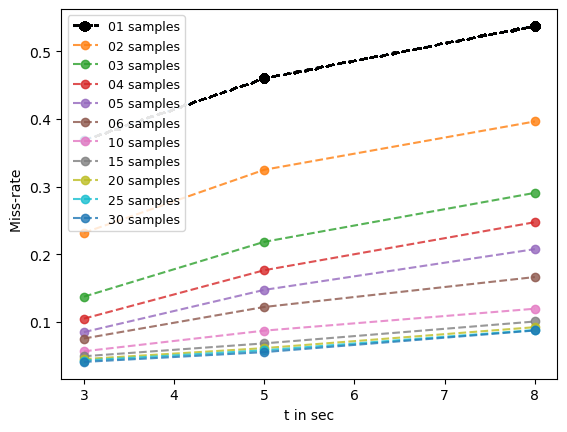}
    \caption{MR comparison with different number of sampled points}
    \label{fig:mr_multiple}
\end{figure}
We show the performance of our approach on sampling $n$ coordinates, $n\in\{1,2,3,4,5,6,10,15,20,25,30\}$. %
The results are shown in \autoref{fig:mr_multiple}. %
With larger $n$, the MR also gets lower, however, the MR does not converge against 0\% MR. %
This can be either because the heatmap does not predict the movement completely, or because the sampling algorithm does not cover the space optimally, e.g., having uncovered gaps between the predictions where the ground truth is. %
Even though this low MR performance can be achieved, it cannot be compared against other approaches because the Waymo motion competition does not measure the performance when predicting more coordinates. %
\subsection{Comparison to the State-of-the-Art}
Our best-performing approach in terms of average SmAP uses the EfficientNet-B0 as the top view encoder and the decoder inspired by HOME\cite{gilles2021HOME} combined with a dynamic grid scaling based on the velocity on the target agent at $t_{0s}$. %
Our best-performing approach is compared against the recent approaches from the literature on the full test dataset, evaluated with the leaderboard. %
The comparison is shown in \autoref{table:missrates_test}. %
Our approach exceeds state-of-the-art performance in terms of MR at $t_{3s}$, second place on average MR, and fourth place for both at $t_{5s}$ and $t_{8s}$. %
For Soft mean average precision (SmAP) our approach scores fifth for $t_{3s}$ and $t_{5s}$, and eighth for $t_{8s}$ and the average. %
All other scores are taken from the leaderboard as of 26.01.2023. %
Compared to the other approaches from the literature, our model performs well on MR for early timesteps. %
However, the performance for later timesteps is worse. %
\section{Discussion}
One of the reasons why our approach is not performing as well for later timesteps might be because the actual threshold of the miss rate is rotated more and therefore, our sampling algorithm does not cover the space as efficiently. %
The rotation of the actual threshold is very large for trajectories like left and right turns ($\approx90$°) and with that our approximation of the threshold area gets worse. %
That could also be one reason why the SmAP placement is worse than the miss rate placement because the trajectories are put into buckets and the final score is an average over all buckets (see \sref{section:metrics}).
\section{Conclusion and Outlook}
The results presented in \sref{chap:Experiments} show that our approach is very competitive in terms of miss rate. %
However, in terms of SmAP, the performance is worse, which could either lead back to worse predicting for a specific type of trajectory or less confidence in the correct predictions. %
A more sophisticated approach to heatmap sampling could increase the performance further, both in terms of miss rate and SmAP. %
It is important to note that from the perspective of an autonomous driving system, good performance for a planning horizon below 3s is paramount. %
This is due to the fact that motion planning operates within this prediction horizon and all prediction errors will lead to irreversible actions that can cause harm. %
Longer prediction horizons influence the strategical behavior planner, which can be used for a smoother driving experience. 

%


\bibliographystyle{./bibtex/IEEEtran} 

\bibliography{./bibtex/IEEEabrv,./bibtex/paper}

\end{document}